\documentclass[sigconf]{acmart}

\setcopyright{acmcopyright}
\copyrightyear{2024}
\acmYear{2024}
\acmDOI{10.48550/arXiv.2408.12003}

\acmConference[International Conference on Artificial Intelligence and Pattern Recognition | AIPR ]{September 20-22, 2024}{Xiamen, China}

\acmISBN{979-8-4007-1717-8}

\acmSubmissionID{XA5000}

\usepackage{algorithmic}
\usepackage{algorithm}
\usepackage{amsmath}

\usepackage{multirow} 
\usepackage{tabularx} 

\begin{document}

\title{RAG-Optimized Tibetan Tourism LLMs: Enhancing Accuracy and Personalization}

\author{Jinhu Qi}
\affiliation{%
  \institution{Chengdu Jincheng College}
  \city{Chengdu}
  \state{Sichuan}
  \country{China}
}
\email{qijinhu1218@gmail.com}

\author{Shuai Yan}
\authornote{Corresponding author}
\affiliation{%
  \institution{Chengdu Jincheng College}
  \city{Chengdu}
  \state{Sichuan}
  \country{China}
}
\email{yanshuai1@cdjcc.edu.cn}

\author{Yibo Zhang}
\affiliation{%
  \institution{Chengdu Jincheng College}
  \city{Chengdu}
  \state{Sichuan}
  \country{China}
}
\email{z1575075389@gmail.com}

\author{Wentao Zhang}
\affiliation{%
  \institution{Chengdu Jincheng College}
  \city{Chengdu}
  \state{Sichuan}
  \country{China}
}
\email{vraniumzwt@gmail.com}

\author{Rong Jin}
\affiliation{%
  \institution{Chengdu Jincheng College}
  \city{Chengdu}
  \state{Sichuan}
  \country{China}
}
\email{kim.rong.king@gmail.com}

\author{Yuwei Hu}
\affiliation{%
  \institution{Chengdu Jincheng College}
  \city{Chengdu}
  \state{Sichuan}
  \country{China}
}
\email{huy791745@gmail.com}

\author{Ke Wang}
\affiliation{%
  \institution{Chengdu Jincheng College}
  \city{Chengdu}
  \state{Sichuan}
  \country{China}
}
\email{wangke@cdjcc.edu.cn}

\begin{abstract}
With the development of the modern social economy, tourism has become an important way to meet people's spiritual needs, bringing development opportunities to the tourism industry. However, existing large language models (LLMs) face challenges in personalized recommendation capabilities and the generation of content that can sometimes produce hallucinations. This study proposes an optimization scheme for Tibet tourism LLMs based on retrieval-augmented generation (RAG) technology. By constructing a database of tourist viewpoints and processing the data using vectorization techniques, we have significantly improved retrieval accuracy. The application of RAG technology effectively addresses the hallucination problem in content generation. The optimized model shows significant improvements in fluency, accuracy, and relevance of content generation. This research demonstrates the potential of RAG technology in the standardization of cultural tourism information and data analysis, providing theoretical and technical support for the development of intelligent cultural tourism service systems.
\end{abstract}



\maketitle

\section{Introduction}
\subsection{Current issues regarding Tibet’s cultural tourism and LLMs}
With the development of modern society's economy, people's desire for spiritual fulfillment is increasing. In terms of tourism, more and more people prefer to visit viewpoints that suit their preferences rather than popular ones. At this point, the abundance of information and inadequate viewpoint recommendations have become significant factors affecting the tourism experience. The Tibet region, with its rich tourism resources but substantial development challenges, urgently needs technological innovation to tap its cultural value and commercial potential.

To address these challenges, this study, under Project ID: XZ2024-01ZY0008, titled "Research and Application of Intelligent Tourism Service System in Tibet Based on LLM," aims to enhance tourists' travel experiences and promote the development of Tibet's tourism industry. The project focuses on the systematic organization and classification of viewpoints in the region. This includes detailed documentation of each viewpoint's historical background, geographical location, cultural significance, and transportation information \cite{Chen2020}.

By constructing a comprehensive database and leveraging contemporary large language models (LLMs), we can offer personalized and precise attraction recommendations based on tourists' needs. Specifically, this approach can guide tourists to less well-known but uniquely charming attractions, thereby alleviating the pressure on popular sites and promoting the rational utilization of tourism resources. Despite the significant technological advancements in LLMs, traditional models in the domain of personalized tourism recommendations still face two critical challenges: insufficient capability in personalized precise recommendations and issues related to the breadth, depth, timeliness, and accuracy of data.

Regarding the insufficiency in personalized recommendation capabilities, traditional models emphasize generalization but lack targeted fine-tuning, leading to suboptimal performance in personalized tourism contexts. Specifically, these models may exhibit "hallucination" phenomena\cite{Chen2023}, where they fail to accurately understand user needs and recommend the most suitable attractions. This limitation is particularly evident in the context of Tibetan tourism, where the unique characteristics of many emerging attractions are not effectively captured, resulting in recommendations that lack accuracy and relevance.

Our research addresses these challenges by employing a vector database query method within the Retrieval-Augmented Generation (RAG) framework\cite{meta_rag_2020}. LLM extracts user requirements, which are then matched against a vector database to identify suitable attractions. These viewpoints are returned to the LLM for integration, resulting in accurate and contextually relevant recommendations tailored to the user's preferences.

The breadth, depth, timeliness, and accuracy of data represent another major challenge affecting the personalized recommendation capabilities of LLMs. As Tibet's tourism resources and infrastructure continue to evolve, LLMs require frequent updates to reflect the latest information. However, existing models often struggle to acquire and integrate the most recent tourism information and user feedback in real-time. This issue is particularly pronounced in remote areas or lesser-known attractions, where detailed data records and descriptions are often insufficient, leading to incomplete and inaccurate recommendations. Moreover, the unique cultural background of Tibet presents challenges for LLMs in handling relevant details with accuracy and depth. For instance, the understanding of Tibetan Buddhist rituals, festivals, and customs may be limited. Although LLMs possess multilingual processing capabilities, they may still face limitations when dealing with the Tibetan language and specialized terms related to Tibetan culture, potentially affecting the accuracy of information transmission.

\section{Methods and Models}

\subsection{Vectorization Methods: TF-IDF and BERT}
TF-IDF (Term Frequency-Inverse Document Frequency) is used to evaluate the importance of a word in a document collection or corpus. By multiplying TF (term frequency) and IDF (inverse document frequency), the TF-IDF value is obtained, which measures the significance of a word to a particular document\cite{Havrlant2017}.

BERT (Bidirectional Encoder Representations from Transformers) handles unlabelled text data by pre-training deep bidirectional representations\cite{Devlin2019}. This pre-training process involves jointly conditioning on both left and right context in all layers, allowing the pre-trained BERT model to be fine-tuned for various tasks by simply adding an output layer. The advantage of this approach lies in its flexibility to adapt to the text data of this project and its capability to create state-of-the-art models for a wide range of NLP tasks.

\subsection{Retrieval Construction Methods}
Retrieval techniques in vector databases can significantly impact the performance and efficiency of systems in large-scale data processing and querying. This project employs the following methods:

Flat (Brute Force Search): The system compares each vector in the database with the query vector, calculates their distance or similarity, and returns the closest vectors\cite{Li2020}.

HNSW (Hierarchical Navigable Small World): This method constructs a layered, small-world graph structure to accelerate query speed. HNSW supports multiple distance metrics, including l2 (Euclidean distance), l1 (Manhattan distance), and dot product\cite{Lin2019}.

IVFFlat (Inverted File Index with Flat Quantization): This technique clusters database vectors into multiple groups. During querying, it only searches the most relevant clusters to reduce computation.

SQ (Scalar Quantization): This method quantizes vector data to reduce storage and computation requirements\cite{Johnson2017}.

IVFSQ (Inverted File Index with Scalar Quantization): This combines the advantages of IVF and SQ by first clustering the data into multiple groups and then quantizing each cluster.

NSG (Navigable Small World Graph): Another graph-based Approximate Nearest Neighbor (ANN) search method, it constructs a navigable small-world graph for fast retrieval\cite{Fu2019}.

LSH (Locality-Sensitive Hashing): This technique accelerates similarity search by mapping similar data into the same hash buckets\cite{Datar2004}.

\subsection{LLMS Function Calling and Retrieval-Augmented Generation}\label{AA}
LLM Function Calling enables large language models to interact with external functions, perform specific tasks, and generate more practical outputs\cite{Gao2023}. Function calling allows large language models to call external functions or services during the text generation process. By leveraging function calling, large language models can decide whether to invoke the corresponding functions based on instructions and return information in a structured format.

Retrieval-augmented generation (RAG) is a method that combines information retrieval and generation models to improve the performance of language models in handling complex queries and generation tasks. The core of the RAG method lies in its ability to combine pre-trained parametric memory with non-parametric external knowledge bases, accessing these external knowledge sources through a differentiable retrieval mechanism. This combination allows RAG models to rely not only on the internally learned knowledge during language generation but also to dynamically incorporate external information, thereby enhancing the relevance and accuracy of the generated content and eliminating model hallucinations\cite{Lewis2020}.

\section{Theoretical Exposition of RAG's Method for Optimizing Hallucination in Large Language Models}
The comprehensive methodology and procedural workflow of this study are depicted in Figure 1 and  Figure 2. This figure provides a visual representation of the sequential steps and critical processes involved in the research, highlighting the key phases and their interconnections. Through this schematic, readers can gain a clearer understanding of the systematic approach adopted in this investigation.
\begin{figure*}[ht!]
\centering
\includegraphics[width=\textwidth]{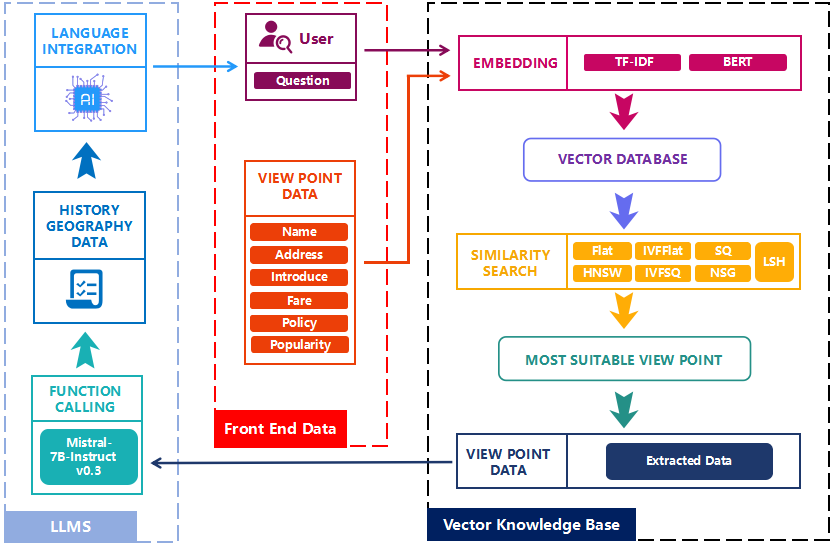} 
\caption{RAG viewpoint information generation system flow chart}
\label{fig:image}
\end{figure*}

\subsection{Dataset construction}
To match the characteristics of each viewpoint when querying with an input prompt and identify the top three viewpoints that best meet the requirements, we integrated and processed existing viewpoint information to create a standardized database. After comparing the accuracy, standardization, and coverage of viewpoint data from various sources in the standardized database, we selected tourism information from Ctrip Travel as the primary data source and supplemented it with information from Wikipedia. For each viewpoint in the database, we recorded key information such as name, province, city, district, address, distance, popularity, ticket prices, brief description, and promotional policies. Through manual selection, we curated a total of 563 viewpoints, saving this information in CSV format as the basis for subsequent experiments.

Additionally, we needed a knowledge source database based on Retrieval-Augmented Generation (RAG) technology to mitigate hallucinations in large-scale language models while retaining crucial information. For this purpose, we employed the open-source large language model Mistral-7B-Instruct-v0.3\cite{Mistral7BInstruct}, which supports function calling. We designed precise prompts and detailed information extraction requirements for the model, embedding them into a function-calling framework. This enabled the large language model to automatically extract required geographic and historical information from Ctrip's tourism information. Extracted information was formatted into a dictionary data structure, containing two primary key-value pairs: one for storing historical information and another for storing geographic information. Ultimately, we successfully extracted and organized the necessary geographic and historical information from the 563 viewpoints corresponding to our standardized database.

\subsection{Database Vectorization and Indexing}
To achieve the task of matching viewpoints based on user preferences and needs while addressing the issue of insufficient retrieval accuracy caused by fuzzy searches, we adopted a method of vectorizing the database before indexing. The classic vectorization methods include TF-IDF and BERT, and the indexing methods for the vectorized data include Flat, HNSWFlat, SQ, LSH, etc. The distance calculation methods after indexing include Manhattan distance, Euclidean distance, and inner product. To explore which combination of "vectorization method + indexing method + distance calculation" is most suitable for the content in the context of Tibet’s cultural tourism, we designed a total of 20 sets of control experiments across these three dimensions.

The prompts for the control experiments were manually created. To ensure randomness in user searches and stability in feature quantity, each prompt randomly includes 1-4 features with an average total feature count of 3. These prompts were used as inputs for each group, and we used the SpaCy library to calculate the number of feature hits and feature hit rates to evaluate the performance of vectorized indexing.

The formula presented below is the formula for TF-IDF\cite{Jones1972}.
\begin{equation}
w_{x,y} = \text{tf}_{x,y} \times \log \left(\frac{N}{\text{df}_x}\right)
\end{equation}

\noindent\textbf{TF-IDF}\\
\noindent\(\text{tf}_{x,y}\) = frequency of \(x\) in \(y\)\\
\(\text{df}_x\) = number of documents containing \(x\)\\
\(N\) = total number of documents
\begin{itemize}
    \item $w_{x,y}$: This represents the TF-IDF weight of term $x$ in document $y$. It is a measure of how important term $x$ is in the specific document $y$.
    \item $\text{tf}_{x,y}$: This is the Term Frequency (TF), which refers to the number of times term $x$ appears in document $y$. It reflects how frequently a word occurs in a document.
    \item $\log \left(\frac{N}{\text{df}_x}\right)$: This part is the Inverse Document Frequency (IDF):
    \begin{itemize}
        \item $N$: The total number of documents in the corpus.
        \item $\text{df}_x$: The number of documents that contain the term $x$.
        \item $\frac{N}{\text{df}_x}$: This ratio represents how common or rare the term $x$ is across all documents in the corpus. The logarithm of this ratio dampens the effect of the term frequency, making sure that common terms (which appear in many documents) do not dominate the weighting.
    \end{itemize}
\end{itemize}

\subsection{RAG Optimization for Large Language Models}
We employed vector database retrieval to address the hallucination problem of large models. After identifying the viewpoints based on user needs, the geographical and historical information from the knowledge source database was returned to the large language model as an external knowledge base. Under this framework, the large language model no longer relies solely on its internal knowledge base to generate viewpoint introduction information. Instead, it understands and integrates the viewpoint introduction content through the external knowledge base. This method not only effectively optimizes the hallucination problem of LLMs but also allows the model to fetch and generate comprehensive viewpoint introduction information, including geographical and historical details, even when certain knowledge is not included within the LLM itself. This research demonstrates the potential of RAG technology in enhancing the accuracy of information generated by large language models, providing new methods and tools for the standardization of tourism information and data analysis.

\subsection{Evaluation Method}
To evaluate the vector databases, we used the keyword-matching function of the SpaCy library along with a mature Chinese pre-trained model. After loading SpaCy's\cite{ExplosionAI} pre-trained Chinese model, we applied it to the result text data from different retrieval methods. By calculating the number of keywords that match the features in the prompts, we can determine the relevance and accuracy of the retrieval results.

Since the dataset used in this experiment is unstructured, we utilized the "calculate composite score" scoring system’s b formula to calculate the scores for various dataset metrics and derive a composite score. Detailed scoring criteria can be referenced from Qi's scoring standards\cite{Qi2024}.

\text{Comprehensive score (percentage system) } = 
\begin{equation}
\begin{cases} 
b & \sum \left(d_1 \cdot \text{fluency } + d_2 \cdot \log(\text{accurate rate} + 1) + \right.\\
  & \left. d_3 \cdot \exp(\text{relevance })\right)
\end{cases}
\end{equation}

For the b formula, we rated the responses of the large language model from three aspects: fluency, accuracy rate, and relevance, and calculated the composite score. The weight coefficients for the different evaluation metrics, d1, d2, and d3, were set with a focus on solving the hallucination problem of the large model. We assigned relevance as the primary analysis metric, with weight parameters set to 0.3, 0.2, and 0.4, respectively. To ensure objectivity in scoring, we chose the fine-tuned Llama3-Chinese model\cite{Wang2024} based on Wang’s work to evaluate the data results of each model. By analyzing the fluency of the generated sentences and the relevance of the generated text to the corresponding knowledge source content, we obtained the scores for fluency and relevance. We used the BERTscore\cite{Zhang2019} evaluation tool developed by Zhang to score the generated text for the accuracy rate. By setting the prompt to control the scores within a range of 0 to 1, with 1 being the highest, we then applied the b formula to calculate the composite score for each model.

\section{Experiment}
\subsection{Database Vectorization and Indexing Experiment}
In constructing and retrieving the vector database, we utilized the widely used FAISS vector library and its associated algorithms (the rationale here may need further justification). For database vectorization, we compared mainstream methods TF-IDF and BERT. For indexing methods, we selected the most representative and characteristic methods: Flat, IVFFlat, NSGFlat, HNSWFlat, SQ, HNSWSQ, IVFSQ, and LSH (the dimension names used here and in Chapter 3 need to be consistent; if necessary, a brief explanation of these concepts should be included at an appropriate place in the text).

Since most of the indexing methods primarily support L2 (Euclidean distance), L1 (Manhattan distance), and Inner-Product (dot product) were only used as comparisons for HNSW in this experiment. Each of the two vectorization methods was tested with the 10 different indexing methods, using the standardized viewpoint dataset established in Section 3.1 for vectorization. We used the 180 prompts constructed following the randomness principle described in Section 3.2 as inputs. To ensure experimental generalization, each prompt randomly included 1 to 4 features, with an average of 3 features per prompt.

The following \textbf{Algorithm 1} outlines a procedure for vectorizing text data (using either TF-IDF or BERT), defining distance metrics and index types, and then querying and saving the results as JSON files. 

\begin{algorithm}
\caption{GenerateIndexResults}
\begin{algorithmic}[1]

\REQUIRE \textit{stop\_words\_file}, \textit{attractions\_csv\_file}
\ENSURE \textit{results\_with\_different\_index\_methods}

\IF{TF-IDF is chosen}
    \STATE vectorizer$\gets$\texttt{TfidfVectorizer(stop\_words, ChineseTokenizer)}
    \STATE text\_vectors $\gets$ vectorizer.fit\_transform(texts).astype(float32)
    \STATE $d \gets$ columns of text\_vectors
\ELSIF{BERT is chosen}
    \STATE model, tokenizer $\gets$ \texttt{BERT('bert-base-chinese')}
    \STATE text\_vectors $\gets$ \texttt{BERT\_encode(texts)}.astype(float32)
    \STATE $d \gets$ size of BERT embeddings
\ENDIF

\STATE distance\_metrics$\gets$\{ 'L2':\texttt{faiss.METRIC\_L2},'L1':\texttt{faiss.METRIC\_L1}, 'Inner Product':\texttt{faiss.METRIC\_INNER\_PRODUCT}\}
\STATE index\_types $\gets$ \{\}

\STATE \textbf{Procedure} \texttt{QueryAndSaveAllAsJSON(index, method\_name, query\_texts, output\_filename)}
    \STATE all\_results $\gets$ \{\}
    \FOR{query\_text \textbf{in} query\_texts}
        \STATE query\_vector$\gets$ 
  vectorizer.transform([query\_text]).astype(float32)
        \STATE $k \gets 3$
        \STATE distances, indices $\gets$ index.search(query\_vector, $k$)
        \FOR{$i \gets 0$ \textbf{to} $k-1$}
            \STATE results\_for\_query.append(\{"description": data.iloc[indices[0][i]]['description'] \})
        \ENDFOR
        \STATE all\_results[query\_text] $\gets$ \{ "query": query\_text, "results": results\_for\_query \}
    \ENDFOR
    \RETURN results\_with\_different\_index\_methods
\end{algorithmic}
\end{algorithm}

\begin{algorithm}
\caption{SpaCy\_avg\_hit\_count}
\begin{algorithmic}[1]

\REQUIRE \textit{json\_files\_directory}, 
\textit{json\_results\_with\_different\_index\_methods}
\ENSURE \textit{average\_scores}

\STATE \textbf{Function} \texttt{AssessAllQueries(json\_path)}
    \STATE queries\_results $\gets$ LoadQueriesResults(json\_path)
    \STATE total\_score, total\_results\_count $\gets$ 0, 0
    \FOR{query\_data \textbf{in} queries\_results.values()}
        \STATE query $\gets$ query\_data["query"]
        \STATE results $\gets$ query\_data["results"]
        \STATE query\_keywords $\gets$ ExtractKeywords(query)
        \FOR{result \textbf{in} results}
            \STATE result\_text $\gets$ Concatenate(result.get("description", ""), result.get("name", ""))
            \STATE result\_keywords $\gets$ ExtractKeywords(result\_text)
            \STATE matched\_keywords $\gets$ Intersection(query\_keywords, result\_keywords)
            \STATE total\_score $\gets$ total\_score + Size(matched\_keywords)
            \STATE total\_results\_count $\gets$ total\_results\_count + 1
        \ENDFOR
    \ENDFOR
    \IF{total\_results\_count > 0}
        \STATE average\_score $\gets$ total\_score / total\_results\_count
    \ELSE
        \STATE average\_score $\gets$ 0
    \ENDIF
    \RETURN average\_score

\end{algorithmic}
\end{algorithm}

This formula represents the percentage improvement of TF-IDF relative to Bert.

Following the methodology outlined in Section 3.2, we retrieved the three closest results for each prompt and confirmed the number of keyword matches using the Chinese model loaded in SpaCy as \textbf{Algorithm 2}. The final results are as follows Table 1.

Calculation of “avg gap”
The formula for calculating the “avg gap” is as follows:
\[
\text{avg gap} = \frac{\text{TF-IDF} - \text{Bert}}{\text{Bert}}
\]

As shown in \textbf{Table 1}, in terms of keyword hit rate, the TF-IDF vectorization method consistently outperformed the BERT vectorization method, with the maximum difference observed in the IVFSQ indexing method. Under the L2 space calculation method, the TF-IDF algorithm's hit rate was 60.1770\% higher than that of the BERT algorithm. The TF-IDF algorithm also had an average hit rate across various indexing methods that was 52.1495\% higher than that of BERT. Additionally, due to differences in the principles of the two vectorization methods, the TF-IDF algorithm reduced vectorization time and performance requirements by approximately 60\% and 47\%, respectively (rough estimates).

Among the TF-IDF vectorization methods, the HNSWFlat method with inner product calculation showed the most outstanding performance, with an average keyword hit rate of 1.9556. However, because the Euclidean distance calculation method is more universal and the difference with the inner product calculation method is less than 1\%, within the experimental error range, considering both universality and performance, we believe that vectorizing the database using the TF-IDF method, constructing the index using the HNSWFlat method, and using L2 Euclidean distance calculation for vector distance is most suitable for this study on Tibetan cultural tourism.

\begin{table*}[h]
\centering
\begin{tabular}{lccccc}
\hline
\textbf{index} & \multicolumn{2}{c}{\textbf{TF-IDF}} & \multicolumn{2}{c}{\textbf{Bert}} & \textbf{avg gap} \\
 & \textbf{avg hit count} & \textbf{avg hit rate} & \textbf{avg hit count} & \textbf{avg hit rate} & \textbf{(bert base)} \\
\hline
Flat & 1.9481 & 64.9383\% & 1.2981 & 43.2716\% & 50.0713\% \\
HNSW\_L2 & \textbf{1.9519} & \textbf{65.0617\%} & 1.2815 & 42.7160\% & 52.3121\% \\
HNSW\_L1 & 1.9463 & 64.8765\% & 1.2796 & 42.6543\% & 52.0984\% \\
HNSW\_IP & \textbf{1.9556} & \textbf{65.1852\%} & 1.2926 & 43.0864\% & 51.2894\% \\
IVFFlat & 1.3500 & 45.0000\% & 0.8370 & 27.9012\% & 61.2832\% \\
SQ & 1.9481 & 64.9383\% & 1.3037 & 43.4568\% & 49.4318\% \\
IVFSQ & 1.3407 & 44.6914\% & 0.8370 & 27.9012\% & 60.1770\% \\
NSG & 1.9500 & 65.0000\% & 1.2981 & 43.2716\% & 50.2140\% \\
LSH & 0.6833 & 22.7778\% & 0.4796 & 15.9877\% & 42.4710\% \\
AVG & 1.6749 & 55.8299\% & 1.1008 & 36.6941\% & 52.1495\% \\
\hline
\end{tabular}
\caption{Comparison of Vectorization and Indexing Methods}
\label{table:1}
\end{table*}
\subsection{Optimization of Large Language Models with RAG Technology}
In our RAG process, we utilized the LLaMA-Factory\cite{Zheng2024}, developed by hiyouga, as the tool for integrating knowledge and making inferences using large language models. We employed this tool with specific parameters to ensure the quality of the language model: a default temperature coefficient of 0.95, a maximum input length of 1024 tokens, and a maximum output length of 512 tokens. Below is the designed instruction for reference: "Please use the provided viewpoint knowledge to introduce the viewpoint to the user. Always adhere strictly to the provided viewpoint information. Input: {query of user, viewpoint information}."

In the research by Jinhu et al., methods such as Supervised Fine-Tuning (SFT) and Optimizing Model Weights (ORPO) were primarily used. These methods generated content about viewpoints under different computation precisions (including BF16, FP16, and FP32) and compared their optimal scores under various large language models, specific computation precisions, and fine-tuning methods. Additionally, these results were compared with the laboratory research on Retrieval-Augmented Generation (RAG) technology. This experiment was conducted using the same large language model, with identical parameter sizes, prompts, and evaluation methods, to scientifically compare the performance differences between RAG technology and different fine-tuning methods in generating viewpoint content while strictly controlling variables. The experimental results indicated significant differences in content quality generated by different fine-tuning methods and computation precisions, while RAG technology demonstrated unique advantages in specific scenarios.

The results of the models optimized with RAG technology are as Table 2.

From the table, we can see that the performance of most experimental models improved after optimization with RAG technology. Keeping the weight parameters consistent and setting the scores of each metric to 1, we obtained a full score of 1.5873 using the formula. The ChatGLM3-6b\cite{huggingface_chatglm3_2023} model achieved a total score of 1.2868 after fine-tuning, and a total score of 1.3228 after RAG optimization, with relevance increasing from 0.7885 to 0.7995, indicating improved model performance. The Baichuan2-7b\cite{huggingface_baichuan2_2023} model had a total score of 1.0012 after fine-tuning, and a total score of 0.9215 after RAG optimization, with relevance decreasing from 0.5485 to 0.4687, suggesting that the model did not perform well after RAG optimization. This issue is likely due to the model's poor performance, making it unsuitable for complex optimization techniques like RAG. The QWEN7b-chat\cite{qwen2_2024} model achieved a total score of 1.1876 after fine-tuning and a total score of 1.3238 after RAG optimization, with relevance increasing from 0.6924 to 0.7966, showing a significant improvement and indicating that RAG technology effectively addresses the hallucination problem in this model. The Llama3-8b model\cite{huggingface_meta_llama_2024} showed the most significant improvement, with relevance increasing from 0.6743 to 0.9721, making the model's performance nearly perfect after optimization. We speculate that the significant impact is due to Llama3-8b's lack of Chinese corpus, preventing effective integration and linking of content from the external knowledge base, resulting in overfitting.

Based on the above experimental results, it is evident that most models perform better in terms of relevance when answering user questions after RAG optimization. Therefore, we believe that RAG technology greatly helps optimize the hallucination problem of large models. Additionally, it can also improve overall model performance metrics such as fluency and accuracy, thereby enhancing the model's generation performance and robustness comprehensively.

\begin{table*}[t]
\centering
\begin{tabular}{l l>{\centering\arraybackslash}p{1.5cm}>{\centering\arraybackslash}p{1.5cm}>{\centering\arraybackslash}p{1.5cm}>{\centering\arraybackslash}p{2cm}>{\centering\arraybackslash}p{2cm}>{\centering\arraybackslash}p{1.5cm}}
\hline
\textbf{Model} & \textbf{Group} & \textbf{Fluency} & \textbf{Accurate Rate} & \textbf{Relevance} & \textbf{Overall Score} & \textbf{Overall Score\%} & \textbf{Improvement} \\
\hline
ChatGLM 3-6b & RAG & 0.8730 & 0.8094 & 0.7995 & 1.3228 & 83.33\% & 0.0279 \\
 & Fine-tuning & 0.8523 & 0.6879 & 0.7885 & 1.2868 & 81.07\% &  \\
Baichuan2-7b & RAG & 0.4435 & 0.6774 & 0.4687 & 0.9215 & 58.05\% & -0.0800 \\
 & Fine-tuning & 0.5734 & 0.6074 & 0.5485 & 1.0012 & 63.08\% &  \\
Qwen-7b-chat & RAG & 0.8876 & 0.8048 & 0.7966 & 1.3238 & 83.40\% & 0.1148 \\
 & Fine-tuning & 0.8018 & 0.6681 & 0.6924 & 1.1876 & 74.82\% &  \\
Llama3-8b & RAG & 0.7530 & 0.8724 & 0.9721 & 1.4643 & 92.25\% & 0.2418 \\
 & Fine-tuning & 0.8150 & 0.6795 & 0.6743 & 1.1792 & 74.29\% &  \\
\hline
\end{tabular}
\caption{Comparison of Large Language Model Performance with RAG Optimization and Fine-Tuning}
\label{table:2}
\end{table*}
\begin{figure*}[ht!]
\centering
\includegraphics[width=2.0\columnwidth]{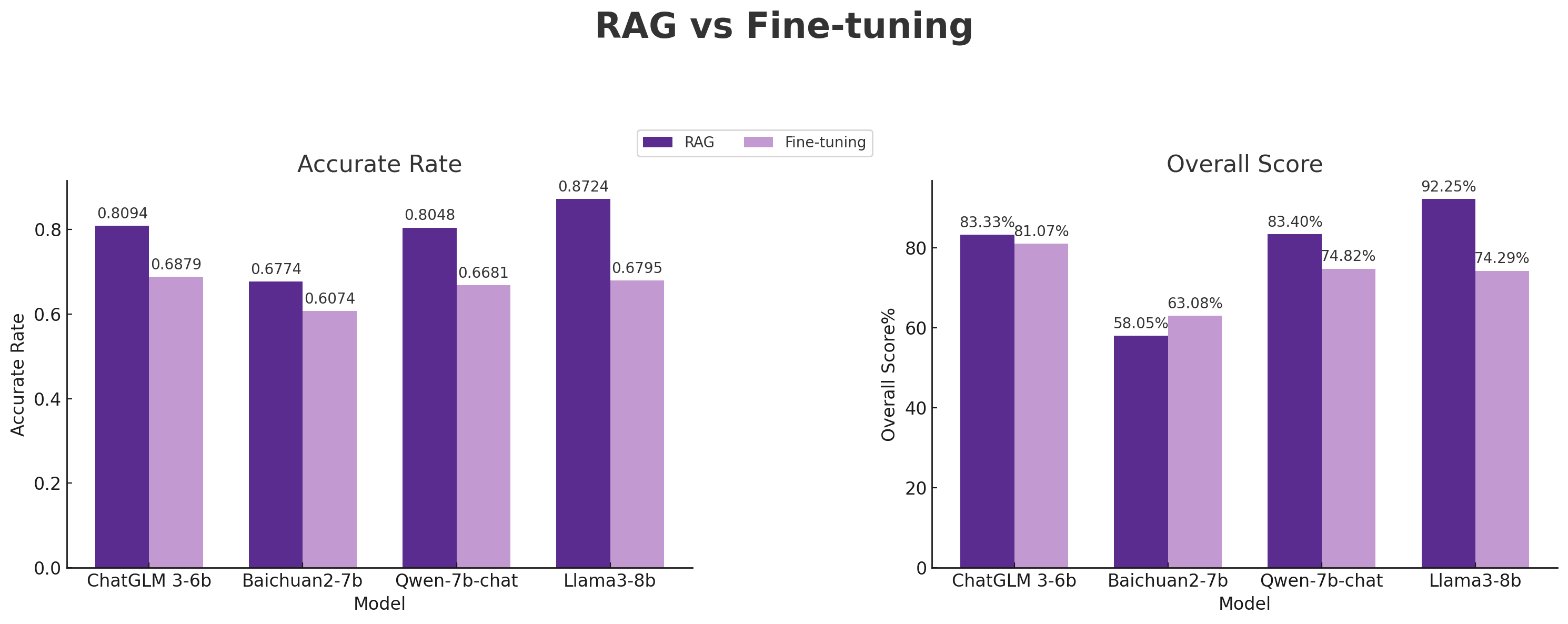} 
\caption{Comparison of the models in RAG and Fine-tuning between Accurate Rate and Overall Score}
\label{fig:image}
\end{figure*}

\section{Conclusion}
This study proposes an optimization scheme based on Retrieval-Augmented Generation (RAG) technology for large language models in the context of Tibet's cultural tourism industry. By analyzing the deficiencies of existing large language models in the application of the cultural tourism field and combining experimental validation, we demonstrated the effectiveness of RAG technology in improving the accuracy of information generation and personalized recommendation capabilities.

First, we integrated existing tourism information resources to construct a detailed database of viewpoints and used vectorization techniques such as TF-IDF and BERT to process the data. Experimental results showed that TF-IDF performed excellently in keyword hit rate and vectorization efficiency. Particularly when combined with the HNSWFlat indexing method and L2 Euclidean distance calculation method, it significantly improved retrieval accuracy. This research provides guidance for the generation and debugging of similar databases in the future.

Second, we applied RAG technology to large language models, utilizing external knowledge bases to address the hallucination problem in content generation. The experimental results indicated that models optimized with RAG showed noticeable improvements in fluency, accuracy, and relevance of content generation. Among them, the Llama3-8b model exhibited the most significant improvement in overall score, increasing from 74.29.\% to 92.25\% after RAG optimization.

Finally, by comparing the performance of various models under different computation precisions and fine-tuning methods, we verified the unique advantages of RAG technology in specific scenarios. This research not only demonstrates the potential of RAG technology in the standardization of cultural tourism information and data analysis but also provides new methods and tools for further optimizing the application of large models in specialized fields.

In conclusion, the vector database construction and model optimization methods employed in this study effectively addressed the challenges of hallucinations and insufficient personalization capabilities in large models for Tibet tourism information recommendation. This provides a solid theoretical and technical foundation for future intelligent cultural tourism service systems. Future research can further explore the application of RAG technology in other fields and how to further enhance the personalized recommendation capabilities and information generation accuracy of large language models.

\begin{acks}
This work is supported by the project “Research and Application of Intelligent Tourism Service System in Tibet Based on LLM” under Grant No. XZ202401ZY0008.
\end{acks}

\end{document}